\documentclass[11pt]{article}
\usepackage{coling2020}
\usepackage{times}
\usepackage{url}
\usepackage{latexsym}

\usepackage{graphicx}
\usepackage{tikz}
\usepackage{pgfplots}
\usepackage{subcaption}
\usepackage{url}
\usepackage{booktabs}
\usepackage{microtype}
\usepackage{placeins}

\newcommand\footnoteref[1]{\protected@xdef\@thefnmark{\ref{#1}}\@footnotemark}

\title{AutoMeTS: The Autocomplete for Medical Text Simplification}

\author{Hoang Van, David Kauchak*, Gondy Leroy\\
  University of Arizona, *Pomona College \\
  {\tt \{vnhh,gondyleroy\}@email.arizona.edu, david.kauchak@pomona.edu}}

\date{}

\begin{document}
\maketitle
\begin{abstract} 
  The goal of text simplification (TS) is to transform difficult text into a version that is easier to understand and more broadly accessible to a wide variety of readers.  In some domains, such as healthcare, fully automated approaches cannot be used since information must be accurately preserved. Instead, semi-automated approaches can be used that assist a human writer in simplifying text faster and at a higher quality.  In this paper, we examine the application of autocomplete to text simplification in the medical domain.  We introduce a new parallel medical data set consisting of aligned English Wikipedia with Simple English Wikipedia sentences and examine the application of pretrained neural language models (PNLMs) on this dataset. We compare four PNLMs (BERT, RoBERTa, XLNet, and GPT-2), and show how the additional context of the sentence to be simplified can be incorporated to achieve better results (6.17\% absolute improvement over the best individual model). 
  We also introduce an ensemble model that combines the four PNLMs and outperforms the best individual model by 2.1\%, resulting in an overall word prediction accuracy of 64.52\%. 

\end{abstract}

\section{Introduction}

The goal of text simplification is to transform text into a variant that is more broadly accessible to a wide variety of readers while preserving the content. While this has been accomplished using a range of approaches \cite{shardlow2014survey}, most text simplification research has focused on fully-automated approaches \cite{xu2016optimizing,zhang2017sentence,nishihara2019controllable}. However, in some domains, such as healthcare, using fully-automated text simplification is not appropriate since it is critical that the important information is preserved fully during the simplification process.  For example, Shardlow et al. \shortcite{shardlow2019neural} found that fully automated approaches omitted 30\% of critical information when used to simplify clinical texts. For these types of domains, instead of fully-automated approaches, interactive text simplification tools are better suited to generate more efficient and higher quality simplifications \cite{kloehn2018jmir}.

Autocomplete tools suggest one or more words during text composition that could follow what has been typed so far. They have been used in a range of applications including web queries \cite{cai2016survey}, database queries \cite{khoussainova2010snipsuggest}, texting \cite{dunlop2000predictive}, e-mail composition \cite{chen2019gmail}, and interactive machine translation,  where a user translating a foreign sentence is given guidance as they type \cite{green-etal-2014-human}. Our work is most similar to interactive machine translation. 
Autocomplete tools can speed up the text simplification process and give full control over information preservation to users, which is required in some domains, such as health and medical.

In this paper, we explore the application of pretrained neural language models (PNLMs) to the autocomplete process for sentence-level medical text simplification. Specifically, given (a) a difficult sentence a user is trying to simplify and (b) the simplification typed so far, the goal is to correctly suggest the next simple word to follow what has been typed. Table \ref{tab:example} shows an example of a difficult sentence along with a simplification that the user has partially typed. An autocomplete model predicts the next word to assist in finishing the simplification, in this case a verb like ``take'', which might be continued to a partial simplification of ``take up glucose''. By suggesting the next word, the autocomplete models provide appropriate guidance while giving full control to human experts in simplifying text.  We explore this task in the health and medical domain where information preservation is a necessity. 

We make three main contributions:

\begin{table}
    \centering
    \scalebox{0.9}{%
    \begin{tabular}{l l}
    \toprule
    Difficult & Lowered glucose levels result both in the reduced release of insulin from the beta \\
              & cells and in the reverse conversion of glycogen to glucose when glucose levels fall. \rule[-2ex]{0pt}{0pt} \\ 
    \hline
    Typed &  This insulin tells the cells to $\rule{1cm}{0.15mm}$ \rule{0pt}{3ex} \\
    \bottomrule
    \end{tabular}}
    \caption{ An example text simplification Autocomplete task.  The user is simplifying the difficult sentence on top and has typed the words on the bottom so far. The example is taken from a medical parallel English Wikipedia sentence pair in Table \ref{tab:medexample}.}
    \label{tab:example}
\end{table}

\begin{table}
    \centering
    \scalebox{0.9}{%
    \begin{tabular}{l l}
    \toprule
    Difficult & Lowered glucose levels result both in the reduced release of insulin from the beta \\
              & cells and in the reverse conversion of glycogen to glucose when glucose levels fall. \rule[-2ex]{0pt}{0pt} \\
    \hline
    Simple & This insulin tells the cells to take up glucose from the blood. \rule{0pt}{3ex}\\
    \bottomrule
    \end{tabular}}
    \caption{An example of sentence pair in Medical Wikipedia parallel corpus.}
    \label{tab:medexample}
\end{table}

\begin{enumerate}

\item We introduce a new parallel medical data set consisting of aligned English Wikipedia and Simple Wikipedia sentences, which is extracted from the commonly used general Wikipedia parallel corpus \cite{kauchak2013improving}. The resulting medical corpus has 3.3k sentence pairs.  This corpus is larger than previously generated corpora (by over 1k sentence pairs) and has stricter quality control \cite{van2019evaluating}. Our corpus requires a medical sentence to contain 4 or more medical words and belong to medical titles as compared to the no title requirement and needing to contain only 1 medical word, as described in Van den Bercken et al. \shortcite{van2019evaluating}.



\item We examine the use of PNLMs for the autocomplete task on sentence-level text simplification and provide an initial analysis based on four recent models on this new medical corpus. In traditional autocomplete tasks, only the text being typed is available.  For text simplification, the additional context of the difficult sentence is also available.  We show how this additional information can be integrated into the models to improve the quality of the suggestions made.

\item We introduce an ensemble model that combines the output of the four PNLMs and outperforms all of the individual models. The ensemble approach is not application specific and may be used in other domains where PNLMs have been employed.

\end{enumerate}

\section{Medical Parallel English Wikipedia Corpus Creation} \label{sec:medical_corpora}

We automatically extracted medical sentence pairs from the sentence-aligned English Wikipedia corpus from Kauchak et al. \shortcite{kauchak2013improving}. Table \ref{tab:medexample} shows an example medical sentence pair. To identify the medical sentence pairs, we first created a medical dictionary with 260k medical terms selected from the Unified Medical Language System (UMLS) \cite{bodenreider2004unified} by selecting all UMLS terms that were associated with the semantic types of: Disease or Syndrome, Clinical Drug, Diagnostic Procedure, and Therapeutic or Preventive Procedure.

We extracted sentences from the English Wikipedia corpus based on the occurrence of terms in the medical dictionary.  Specifically, a sentence pair was identified as medical
if both the title of the article and the English Wikipedia sentence had 4 or more terms that matched the medical keywords. A term was considered a match to the UMLS dictionary if it had a similarity score higher than 0.85 using QuickUMLS \cite{soldaini2016quickumls}. We then manually reviewed the sentence pairs and removed all non-medical sentence pairs. The final medical parallel corpus has 3.3k aligned sentence pairs\footnote{\url{https://github.com/vanh17/MedTextSimplifier/tree/master/data_processing/data}}.



Van den Bercken et al. \shortcite{van2019evaluating} also created a parallel medical corpus by filtering sentence pairs from Wikipedia.  Our corpus is significantly larger (45\% larger; 2,267 pairs vs. 3,300 pairs) and uses a stricter criteria for identifying sentences: they only required a single word match in the text itself (not the title) and used a lower similarity threshold of 0.75 (vs. our approach of 0.85).



\section{Approach}
We had three goals for our analysis: understand the effect of incorporating the additional context of the difficult sentence into autocomplete text simplification models, explore a new application for PNLMs, and evaluate our new ensemble approach to the autocomplete text simplification. 

\subsection{Autocomplete Approach For Medical Text Simplification}
We pose the autocomplete text simplification problem as a language modeling problem: given a difficult sentence that a user is trying to simplify, $d_1 d_2 ... d_m$, and the simplification typed so far, $s_1 s_2 ... s_i$, the autocomplete task is to suggest word $s_{i+1}$.  Table \ref{tab:example} gives an example of the autocomplete task. To evaluate the models, we calculated how well the models predicted the next word in a test sentence, given the previous words.  A simple test sentence of length $n$, $s_1 s_2 ... s_n$, would result in $n-1$ predictions, i.e., predict $s_2$ given $s_1$, predict $s_3$ given $s_1 s_2$, etc.  For example, Table \ref{tab:medexample} shows a difficult sentence from English Wikipedia and the corresponding simplification from the medical Simple English Wikipedia. Given this test example, we generate 12 prediction tasks, one for each word in the simple sentence after the first word.  Table \ref{fig:testing} shows these 12 test prediction tasks.

\subsection{Increasing Information Through Incorporation of Additional Context}

Unlike other autocomplete tasks, for text simplification, the difficult sentence provides very explicit guidance about what words and information should be included as the user types. As a baseline, we compare the models without the context of this additional information, i.e., we predict word $s_{i+1}$ given only the simplification typed so far, $s_1 s_2 ... s_i$.  We take a straightforward approach to incorporate the context of the difficult sentence: we concatenate the difficult sentence in front of a simplification typed so far, i.e., predict $s_{i+1}$ given $d_1 d_2 ... d_m . s_1 s_2 ... s_i$.  This has the advantage of incorporating the difficult sentence and biasing the predictions towards those found in the encoded context from difficult sentences, but is still model-agnostic, allowing us to apply it to all the different PNLMs without modifying the underlying architecture.

\begin{table}
\centering
\scalebox{0.9}{%
\begin{tabular}{c l l}
\toprule
Difficult sentence & Lowered glucose levels result both in the reduced release of insulin & \\
 & from the betacells and in the reverse conversion of glycogen to   & \\
 & glucose when glucose levels fall. & \\
\midrule
\toprule
Prediction Task & Simplification typed so far & Predict \\
\midrule
1 & This & insulin \\
2 & This insulin & tells \\
3 & This insulin tells & the \\
4 & This insulin tells the & cells \\
5 & This insulin tells the cells & to \\
6 & This insulin tells the cells to & take \\
7 & This insulin tells the cells to take & up \\
8 & This insulin tells the cells to take up & glucose \\
9 & This insulin tells the cells to take up glucose & from \\
10 & This insulin tells the cells to take up glucose from & the \\
11 & This insulin tells the cells to take up glucose from the & blood \\
12 & This insulin tells the cells to take up glucose from the blood & .\\
\bottomrule
\end{tabular}}
\caption{The resulting prediction tasks that are generated from the example in Table \ref{tab:medexample}.}
\label{fig:testing}
\end{table}

\subsection{Transformer-based Language Models}

We examined four PNLMs based on the Transformer network \cite{vaswani2017attention}: BERT \cite{devlin2018bert},  RoBERTa \cite{liu2019roberta},  XLNet \cite{yang2019xlnet},  and GPT-2 \cite{radford2019language}.  For each of the models, we examine versions that only utilize the text typed so far, denoted ``No Context'', as well as ``context-aware'' variants that utilize the difficult sentence. We fine-tuned all four models on the 160k sentence pair general parallel English Wikipedia \cite{kauchak2013improving} (excluding the development and testing data) and then further fine-tuned them on the separate medical training set described in section \ref{sec:medical_corpora}.
Note that none of the test sentences were in a dataset used for fine-tuning.

\paragraph{BERT:} Bidirectional Encoder Representations from Transformers  \cite{devlin2018bert} is a method for learning language representations using bidirectional training. BERT has been shown to produce state-of-the-art results in a wide range of generation and classification applications. We use the base original BERT\footnote{\url{https://github.com/huggingface/bert/}} model pre-trained on the BooksCorpus \cite{zhu2015aligning} and English Wikipedia.

\paragraph{RoBERTa:} A Robustly Optimized BERT Pretraining Approach \cite{liu2019roberta}. The RoBERTa uses the same model architecture as BERT. However, the differences between RoBERTa and BERT are that RoBERTa does not use Next Sentence Prediction during pre-training and RoBERTa uses larger mini-batch size. We used the publicly released base RoBERTa\footnote{\url{https://github.com/huggingface/roberta}} with 125M parameters model.

\paragraph{XLNet:} Generalized Auto-regressive Pretraining Method \cite{yang2019xlnet}. Like BERT, XLNet benefits from bidirectional contexts. However, XLNet does not suffer limitations of BERT because of its auto-regressive formulation. In this work, we used publicly available base English XLNet\footnote{\url{https://github.com/huggingface/xlnet}} with 110M parameters model.

\paragraph{GPT-2:} Generative Pretrained Transformer 2 \cite{radford2019language}. Like BERT, GPT-2 is also based on the Transformer network, however, GPT-2 uses unidirectional left-to-right pre-training process. We use the publicly released GPT-2\footnote{\url{https://github.com/huggingface/gpt2}} model, which has 117M parameters and is trained on web text.

\subsection{Ensemble Models} \label{sec:ensemble}

Each of the models above utilizes different network variations and was pretrained on different datasets, and therefore they do not always make the same predictions.  Ensemble approaches combine the output of different systems to try and leverage these differences to create a single model that outperforms any of the individual models.  We examined three ensemble approaches that combine the output of the four models.

\paragraph{Majority Vote:} As a baseline ensemble approach, we examined a simple majority vote on what the next word should be. We take the top 5 suggestions from each of the models and do a majority count on the pool of combined suggestions.  The output of the model is the suggestion with highest count. If there is a tie, we randomly select one of the top suggestions. We picked the top 5 suggestions since this was the cutoff where the models tended to peak on the development data (for example, see the accuracy@N as shown in Table \ref{tab:accuracy-@-n}. Having more than 5 suggestions did not improve performance much but slowed the computation.

\paragraph{4-Class Classification (4CC):} \label{sec:4cc} The ensemble problem can be viewed as a classification problem where the goal is to predict which system output should be used given a difficult sentence and the words typed so far, i.e., the autocomplete example. We posed this as a supervised classification problem.  Given an autocomplete text simplification example, we can generate training data for the classifier by comparing the output of each system to the correct answer. If a system does get the example correct, then we include an example with that system as the label. Table \ref{tab:4ccexample} shows three such examples, where RoBERTa correctly predicted the first example, BERT the second, and XLNet the third.  If multiple systems get the example right, we then randomly assign the label to one of the systems.

We train a neural text classifier implemented by huggingface\footnote{\url{https://github.com/huggingface/transformers}} with this training set to the PNLMs given the next-word prediction task. To make use of the models' confidence on top of the results from model selection, we designed a scoring system for output selection as follows:
\[ \label{eq:4cc}
	Score(w, X) = \alpha * P(w|X) + \theta * I (X, S)
\]

\noindent where $P(w|X)$ is model $X$'s confidence on predicted word $w$; $I(X, S)$ is an identity function, which returns $1$ if $X = S$ and $0$ otherwise; $S$ is the predicted model from model selector; and $\alpha$ and $\theta$ are scoring parameters. We use $0.5$ for both $\alpha$ and $\theta$.  
At testing time, we pick the highest score and output the word $w$, given a prediction task.
\begin{table}
    \centering
    \scalebox{0.9}{%
    \begin{tabular}{l l}
    \toprule
    Prediction Task  & Class \\
    \midrule
    (Difficult sentence). This (MASK) & RoBERTa \\
    (Difficult sentence). This insulin (MASK) & BERT \\
    (Difficult sentence). This insulin tells (MASK) & XLNet \\
    \bottomrule
    \end{tabular}}
    \caption{An example of training data for the 4CC model. Class can be one of the four option: BERT, RoBERTa,  XLNet, GPT-2.}
    \label{tab:4ccexample}
    \
\end{table}

\paragraph{Autocomplete for Medical Text Simplification (AutoMeTS):} RoBERTa performed significantly better than the other three individual models at the simplification autocomplete task. As a result, there was a strong bias toward RoBERTa in the training data for the 4CC ensemble model.  To mitigate this effect, we also developed an ensemble approach based on a multi-label classifier for model selector, which we denote the AutoMeTS ensemble model. This choice of model selector, to our knowledge, is novel to transformer-based ensemble models. For this choice of classifier, each prediction task is given a sequence of 4 binary labels. Each label represents the correctness of each of the individual PNLMs, with a 0 representing an incorrect prediction on the task and a 1 representing a correct prediction. Table \ref{tab:AutoMeTSexample} shows an example of this dataset with the labels in order ``RoBERTa BERT XLNet GPT-2''.  For the first example, RoBERTa, XNLET, and GPT-2 correctly predicted the next word, while BERT did not.

We trained a neural multi-label classifier implemented by huggingface on this training dataset. To make use of the models' confidence on top of the results from model selection, we designed a scoring system for output selection as follows:
\[ \label{eq:AutoMeTS}
	Score(w, X) = \beta * P(w|X) + \sigma * S(X, Ls)
\]

\noindent where $P(w|X)$ is model $X$'s confidence on predicted word $w$; $S(X, Ls)$ is a function, which returns $0.25$ if model $X$ is in $Ls$ and $0$ otherwise; $Ls$ is the predicted sequence of labels from the model selector; and $\beta$ and $\sigma$ are scoring parameters. We use $0.5$ for both $\beta$ and $\sigma$. At testing time, we output the word $w$ with the highest score, given a prediction task.

\begin{table}
    \centering
    \scalebox{0.9}{%
    \begin{tabular}{l c}
    \toprule
    Prediction Task  & Sequence of Labels \\
    \midrule
    (Difficult sentence). This (MASK) & 1 0 1 1 \\
    (Difficult sentence). This insulin (MASK) & 0 1 0 0 \\
    (Difficult sentence). This insulin tells (MASK) & 1 1 1 1 \\
    \bottomrule
    \end{tabular}}
    \caption{An example of training data for the AutoMeTS model. For a prediction task, a sequence of 4 labels is give in the order "RoBERTa BERT XLNet GPT-2". The value of 1 means the model correctly predicted the right word, and 0 otherwise.}
    \label{tab:AutoMeTSexample}
    \
\end{table}

\section{Experimental Setup} \label{sec:results}

We compare the performance of the models on the medical autocomplete text simplification task.  We used  our medical parallel English Wikipedia corpus with 70\% of the sentence pairs for training, 15\% for development, and 15\% for testing. We fine-tuned individual PNLMs using huggingface\footnote{\url{https://github.com/huggingface/}} with a batch-size of 8, 8 epochs, and a learning rate of $5e^{-5}$. Early stopping was used based on the second time a decrease in the accuracy was seen.


We used two metrics to evaluate the quality of the approaches.  First, we used standard accuracy, where a prediction is counted correct if it matches the test prediction word. Accuracy is pessimistic in that the predicted word must match exactly the word seen in the simple sentence, and as such it does not account for other possible words, such as synonyms, that could be correctly used in the context. Since the parallel English Wikipedia corpus does not offer multiple simplified versions for a given difficult sentence, accuracy is the best metric that considers automated scoring, simplification quality, and information preservation. Accuracy-based metrics can help offset an expensive manual evaluation while providing the best approximation of how the autocomplete systems work. We do not use BLEU \cite{papineni2002bleu} and SARI \cite{xu2016optimizing} scores, which are widely used in text simplification domain, because the two metrics are specifically designed for fully-automated models that predict an entire sentence at a time.  For autocomplete, the models only predict a single word at a time and then, regardless of whether the answer is correct or not, use the additional context of the word that the user typed next to make the next prediction. 

Autocomplete models can suggest just the next word, or they can be used to suggest a list of alternative words that the user could select from (since the models are probabilistic they can return a ranked list of suggestions).  To evaluate this use case, and to better understand what words the models are predicting, we also evaluated the models using accuracy@N. Accuracy@N counts a model as correct for an example as long as it suggests the correct word within the first $N$ suggestions.

\begin{figure}[t]
\center{\includegraphics[scale=0.7]
{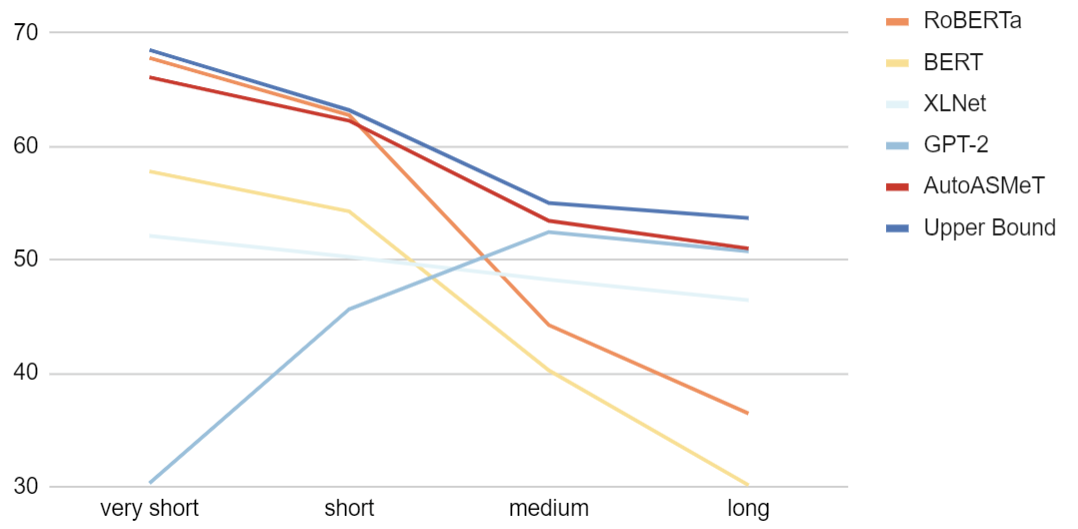}}
\caption{\label{fig:length_based} Accuracy for the context-aware models based on the length of the difficult sentences: very short ($\le 5$ tokens), short ($6-15$), medium ($16-19$), and long ($\ge20$).}
\end{figure}

\section{Results and Discussion}

We first analyze the results of the individual PNLMs on the medical text simplification autocompletion task and then explore the ensemble approaches.  We also include a number of post-hoc analyses to better understand what the different models are doing and limitations of the models.

\subsection{Individual Language Models}


\paragraph{Accuracy} Table \ref{tab:results} shows the results for the four different variants (RoBERTa, BERT, XLNet, and GPT-2 with and without context). Even without any context, many of the models get every other word correct (accuracy of around 50\%).  With the additional context of the difficult sentence, all models improve.  GPT-2 improves drastically (more than doubling the accuracy) and RoBERTa also achieves a reasonable improvement of 6\% absolute.  Both with and without context, RoBERTa is the best performing model achieving significantly higher results than the other models


\begin{table}[t]
\centering
\scalebox{0.9}{%
\begin{tabular}{l c c }
\toprule
Model & No Context & Context-Aware\\
\midrule
Single PNLMs & & \\
\midrule
RoBERTa & \textbf{56.23} & \textbf{62.40} \\
BERT & 50.43 & 53.28 \\
XLNet & 45.70 & 46.20 \\
GPT-2 & 23.2 & 49.00 \\
\midrule
Ensemble Models & & \\
\midrule
Majority Vote & 39.75 & 48.25 \\
4CC & 52.27 & 59.32 \\
AutoMeTS & \textbf{57.89} & \textbf{64.52} \\
Upper bound & 60.22 & 66.44 \\
\bottomrule
\end{tabular}}
\caption{Accuracy of pretrained neural language models (PNLMs), both with and without the context of the difficult sentence on the medical parallel English Wikipedia corpus.} 
\label{tab:results}
\
\end{table}

\begin{table}[t]
\centering
\scalebox{0.9}{%
\begin{tabular}{l c c c c}
\toprule
 & RoBERTa & BERT & XLNet & GPT-2\\
 \midrule
accuracy@1 & 62.40 & 53.28 & 46.20 & 49.00 \\
accuracy@2 & 67.20 & 54.50 & 46.90 & 49.44\\
accuracy@3 & 70.00 & 56.20 & 49.20  & 52.57\\
accuracy@4 & 72.10 & 58.00 & 51.30 & 54.32\\
accuracy@5 & \textbf{73.20} & \textbf{59.40} & \textbf{53.50} & \textbf{56.12}\\
 \bottomrule
\end{tabular}}
\caption{\label{tab:accuracy-@-n} Accuracy@N of the RoBERTa, BERT, XLNet, and GPT-2 with context on next word prediction.}
\vspace{-2mm}
\end{table}

\paragraph{Accuracy@N} Table \ref{tab:accuracy-@-n} shows the accuracy@N from PNLMs on next word prediction.  By allowing the autocomplete system user the option to pick from a list of options, the correct word is much more likely to be available.  Even just showing three options, results in large improvements, e.g., 7.5\% absolute for RoBERTa.  When five options are available, increases range from  6--10.8\%.

\paragraph{Impact of the difficult sentence length} To better understand the models, we compared the average performance of the models based on the length of the sentence that was being simplified.  We divided the test sentence into four different groups based on length: very short ($\leq$5 tokens), short ($6-15$ tokens), medium ($16-19$ tokens), and long ($\geq$20 tokens).  Figure \ref{fig:length_based} shows the test accuracy of the context-aware models broken down into these 4 different groups. RoBERTa, BERT, and XLNet are fairly consistent regardless of the difficult sentence lengths; only for long sentences does the performance drop. GPT-2 performs poorly on short sentences, but well for other lengths. We hypothesize that the training data for GPT-2 (web text) may require more context for this more technical task.

\begin{figure}[t]
\center{\includegraphics[scale=0.7]
{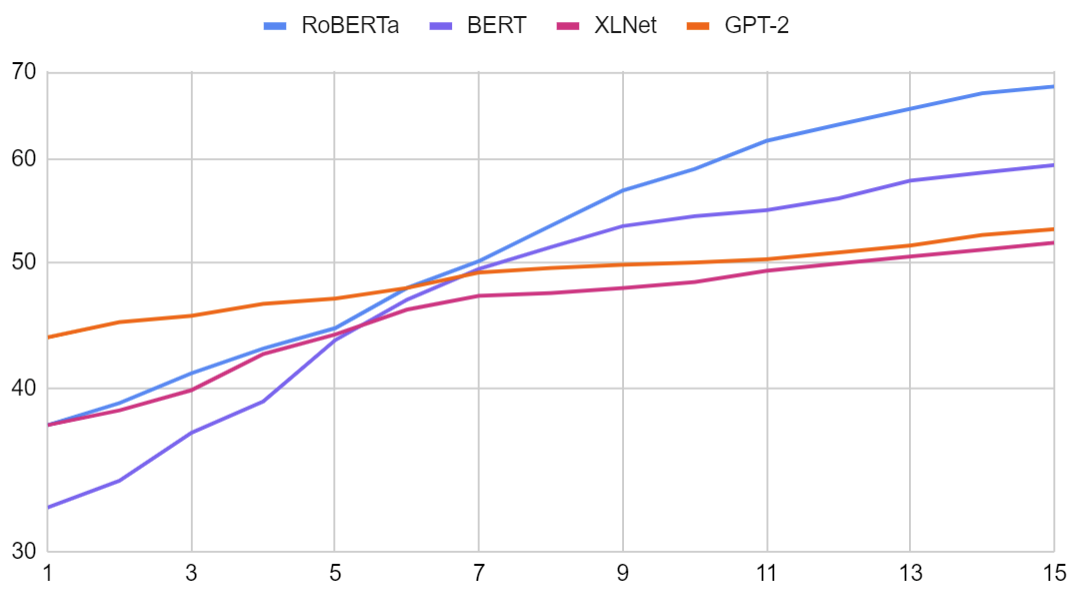}}
\caption{Accuracy for the four context-aware models based on the number of words typed so far.}
\label{fig:position_based}
\end{figure}

\paragraph{Impact of the number of words typed}  We also analyzed the average performance of the models based on how many of the words of the simplified sentence had been typed so far, i.e., the length of $s_1 s_2 ... s_i$.  Figure \ref{fig:position_based} shows the performance accuracy of the models based on how many words of the simplification the model has access to. Early on, when the sentence is first being constructed, all models perform poorly. As more words are typed, the accuracy of all models increases.  GPT-2 performs the best early on, but then levels off after about 7 words.  Both BERT and RoBERTa continue to improve as more context is added, which may partially explain their better performance overall.

\begin{figure}[t]
\center{\includegraphics[scale=0.7]
{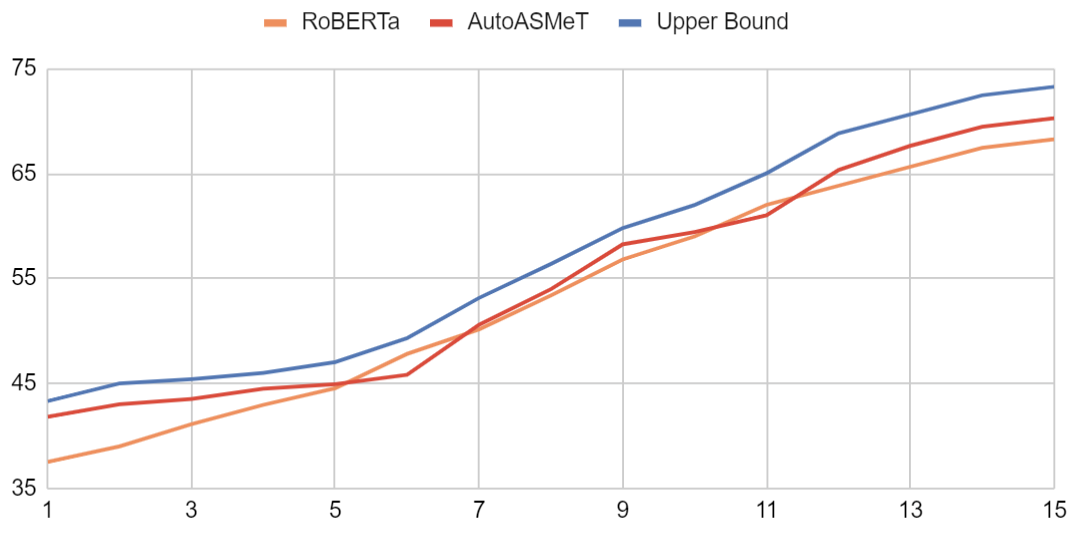}}
\caption{\label{fig:ensemble_pos} Accuracy for the RoBERTa, AutoMeTS, and upper bound models based on the number of words typed so far.}
\end{figure} 

\subsection{Ensemble Models}

Although RoBERTa performs the best overall, as Figures \ref{fig:length_based} and \ref{fig:position_based} show, different models perform better in different scenarios.  The ensemble models try and leverage these differences to generate a better overall model.

As shown in Table \ref{tab:results}, the majority vote ensemble model does not perform better than the best individual PNLM.  The 4CC does outperform the majority vote approach by 11.07\% and does perform better than three of the four PLNMs, but it still fails to beat RoBERTa.  AutoMeTS, by viewing the problem as a multi-label problem, is able to avoid some of the biases in the training data that 4CC has, resulting in an absolute improvement of 2.1\% over the best individual model (RoBERTa).

To understand the performance differences of the ensemble models,  Table \ref{tab:model_frequency} shows the percentage that each of the four PNLMs was used for each of the ensemble approaches.  The problem with the majority vote is that it tends to utilize all of the systems, regardless of their quality.  For example, it shows a high percentage of XLNet, even though its performance was the worst.  Because RoBERTa is the best performing model, the 4CC approach had a very strong bias towards RoBERTa. The multi-label selector reduces the bias towards using RoBERTa (a 11.25\% decrease in the appearance of RoBERTa) and is able to leverage predicions from the other model when appropriate.

\begin{table}[t]
\centering
\scalebox{0.9}{%
\begin{tabular}{l c c c}
\toprule
	 	&	Majority Vote	&	4CC & AutoMeTS\\
\midrule
RoBERTa & 47.29\%  & 	71.00\% 		& 	59.75\% \\
BERT 	& 20.25\%  & 	12.45\% 		& 	18.09\% \\
XLNet 	& 15.41\%  & 	5.72\% 		& 	7.06\% \\
GPT-2 	& 17.05\%  & 	10.83\% 		& 	15.10\% \\
\bottomrule
\end{tabular}}
\caption{The appearance frequency of PNLMs in Majority Vote, 4CC, AutoMeTS ensemble models.}
\label{tab:model_frequency}
\vspace{-2mm}
\end{table}

To understand the limits of an ensemble  approach, we also calculated the upper bound that the ensemble approach could achieve.  Specifically, as long as at least one model among the four PNLMs correctly predicts the next word, we mark it as correct for the upper bound. This means that no other possible combination of the four PNLMs can perform better. Here this upper bound is 66.44\% (Table \ref{tab:results}), which is about a 2\% improvement over our ensemble approach; there is a bit of room for improvement, but also better language modeling techniques also need to be explored.  

Figure \ref{fig:ensemble_pos} shows the accuracy for RoBERTa, AutoMeTS, and the upper bound based on size of the context, i.e., words typed so far. For small context, the ensemble approach performs much better than RoBERTa.  This likely can be attributed to selecting one of the other models that performs better for small context, e.g., GPT-2.  As the context size increases, however, RoBERTa and the ensemble model perform similarly.  The upper bound is consistently above the ensemble approach across all context sizes.

\section{Conclusions}

In this paper, we introduced a new medical parallel English Wikipedia corpus for text simplification, which contains 3.3K medical sentence pairs. Further, we proposed a new autocomplete application for PNLMs for medical text simplification. Such autocomplete models can assist users in simplifying text with improved efficiency and higher quality results in domains where information preservation is especially critical, such as healthcare and medicine, and where fully-automated approaches are not appropriate. We examined four recent PNLMs: BERT, RoBERTa, XLNet, and GPT-2, and showed how the additional context of the sentences being simplified could be incorporated into the autocomplete simplification process. Further, we introduced AutoMeTS, an ensemble method that combines the advantages of each of the different PNLMs. The AutoMeTS model outperforms the best individual model, RoBERTa, by 2.1\%. Longer term, we envision that this new application could lead to other interesting model adaptations and advance text simplification in medical domains.

\section*{Acknowledgements}

Research reported in this paper was supported by the National Library of Medicine of the National Institutes of Health under Award Number R01LM011975. The content is solely the responsibility of the authors and does not necessarily represent the official views of the National Institutes of Health.


\bibliographystyle{coling}
\bibliography{coling2020}

\end{document}